\documentclass[sigconf]{acmart}

\renewcommand\footnotetextcopyrightpermission[1]{}
\setcopyright{none}

\begin{document}

\title{ID-Eraser: Proactive Defense Against Face Swapping via Identity Perturbation}

\author{Junyan Luo}
\email{ShaoyeXiaoye@163.com}
\affiliation{%
  \institution{Jinan University}
  \city{Guangzhou}
  \country{China}
}

\author{Peipeng Yu}
\email{ypp865@163.com}
\orcid{0000-0003-0056-4300}
\affiliation{%
  \institution{Jinan University}
  \city{Guangzhou}
  \country{China}
}

\author{Jianwei Fei}
\email{fjw826244895@163.com}
\affiliation{%
  \institution{University of Florence}
  \city{Florence}
  \country{Italy}
}

\author{Zhihua Xia}
\email{xia_zhihua@163.com}
\authornote{Corresponding author: xia\_zhihua@163.com.}
\affiliation{%
  \institution{Jinan University}
  \city{Guangzhou}
  \country{China}
}

\author{Shiya Zeng}
\email{zengshiya@stu2024.jnu.edu.cn}
\affiliation{%
  \institution{Jinan University}
  \city{Guangzhou}
  \country{China}
}

\author{Xiaoyu Zhou}
\email{xiaoyuzhou68@outlook.com}
\affiliation{%
  \institution{Jinan University}
  \city{Guangzhou}
  \country{China}
}

\author{Xiang Liu}
\email{liuxiang@dgut.edu.cn}
\affiliation{%
  \institution{Dongguan University of Technology}
  \city{Dongguan}
  \country{China}
}

\renewcommand{\shortauthors}{Luo et al.}

\begin{abstract}
Deepfake technologies have rapidly advanced with modern generative AI, and face swapping in particular poses serious threats to privacy and digital security. Existing proactive defenses mostly rely on pixel-level perturbations, which are ineffective against contemporary swapping models that extract robust high-level identity embeddings.
We propose ID-Eraser, a feature-space proactive defense that removes identifiable facial information to prevent malicious face swapping. By injecting learnable perturbations into identity embeddings and reconstructing natural-looking protection images through a Face Revive Generator (FRG), ID-Eraser produces visually realistic results for humans while rendering the protected identities unusable for Deepfake models.
Experiments show that ID-Eraser substantially disrupts identity recognition across diverse face recognition and swapping systems under strict black-box settings, achieving the lowest Top-1 accuracy (0.30) with the best FID (1.64) and LPIPS (0.020). Compared with swaps generated from clean inputs, the identity similarity of protected swaps drops sharply to an average of 0.504 across five representative face swapping models. ID-Eraser further demonstrates strong cross-dataset generalization, robustness to common distortions, and practical effectiveness on commercial APIs, reducing Tencent API similarity from 0.76 to 0.36.
\end{abstract}




\maketitle
\thispagestyle{plain}

\begin{figure}[h]
  \centering
  \includegraphics[width=\linewidth,trim=1 1 1 1,clip]{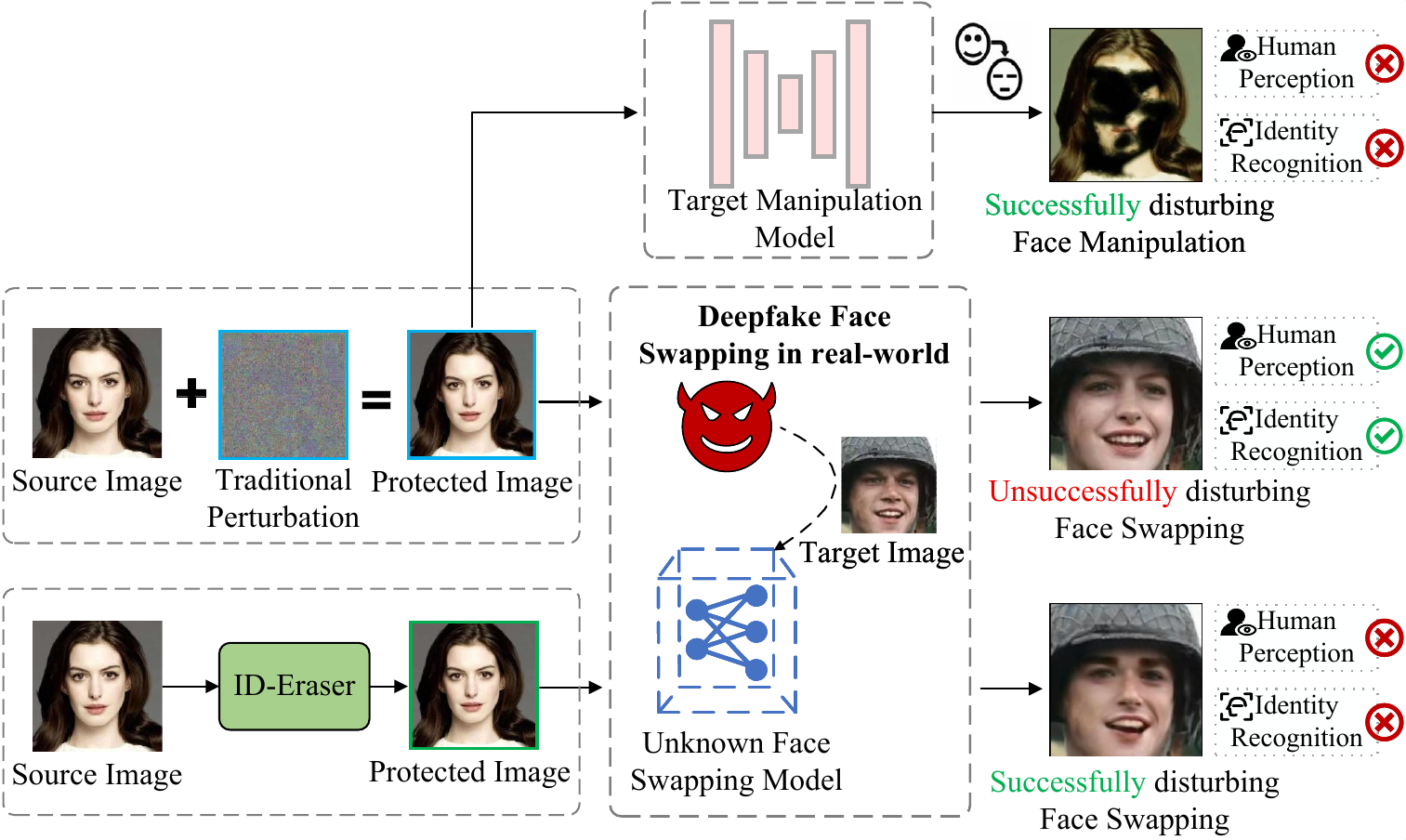} % 推荐使用\linewidth控制宽度
  \caption{Pixel-level defenses perturb the target image, while ID-Eraser protects the 
source identity via feature-space perturbation and remains effective against 
unknown face swapping models.}
  \label{fig:pic1}
\end{figure}
\section{Introduction}
\label{sec:introduction}

The rapid development of generative AI has greatly accelerated the advancement of Deepfake technologies. 
While these technologies offer positive applications in entertainment and creative industries, their misuse—particularly face swapping—has raised severe security and privacy concerns. 
Face swapping transfers a victim’s facial identity onto a target image while retaining the target’s expression, pose, and background, enabling the creation of realistic forged content that can spread misinformation, damage reputations, and even bypass liveness detection systems.

To mitigate the societal risks posed by Deepfakes, prior studies follow two major directions: Deepfake detection and proactive protection. 
Detection methods~\cite{rethinking,laa,speechforensics,c2p,fish,vigo} perform post-hoc forensic analysis but do not prevent identity misuse. 
Furthermore, Li et al.~\cite{seeing} demonstrate that face swapping can easily bypass many facial liveness verification systems, highlighting the need for proactive protection.

Proactive defenses aim to modify images before manipulation occurs. 
Most existing approaches~\cite{id-guard,initiative,DF-rap,anti,upwd} add imperceptible perturbations to the input image to degrade the output quality of specific manipulation models. 
However, these defenses are largely designed for attribute editing or facial reenactment, rather than for true face swapping attacks. 
Their perturbations are typically model-specific and fail in realistic black-box settings. 
Universal-perturbation methods~\cite{visually,restricted} also protect the target image rather than the source identity, and their perturbations typically cause only blurring or artifacts without effectively disrupting identity features. As a result, the swapped faces may still preserve the victim’s identity. Crucially, while these defenses aim to protect the target victim image, face swapping creates harm by exploiting the source identity. 
In other words, once an attacker obtains the victim's facial image, it can be swapped onto any target, regardless of whether that target image has been protected. 
Therefore, existing defenses are fundamentally insufficient for preventing Deepfake face swapping, as illustrated in figure~\ref{fig:pic1}.

To address these challenges, we propose \textbf{ID-Eraser}, the first feature-level proactive defense against Deepfake face swapping based on identity erasure. 
Unlike pixel-level defenses, ID-Eraser perturbs the latent identity representation extracted from the source image and reconstructs a natural-looking protected image from the perturbed features. 
The resulting images appear unchanged to human observers but prevent face swapping models from extracting valid identity embeddings, causing the generated swapped faces to fail in reproducing the victim's identity.

Specifically, for a given input face, we first use a pre-trained identity extractor to obtain its identity features. These features are then passed through the Feature Perturbation Module (FPM) to generate a perturbed identity vector. Next, the Face Revive Generator (FRG) reconstructs an image that is visually indistinguishable from the original image, thereby embedding the protection while preserving appearance consistency. Our main contributions are summarized as follows:
\begin{itemize}
    \item We propose ID-Eraser, a proactive defense that introduces a 
    feature-level identity perturbation framework for Deepfake face swapping. 
    Unlike prior pixel-level defenses, ID-Eraser requires no surrogate model or 
    target-model gradients and remains fully applicable in restricted black-box scenarios.

    \item We introduce a feature-level protection framework that perturbs identity embeddings through a dedicated Feature Perturbation Module (FPM), and reconstructs natural-looking protected images using a Face Revive Generator (FRG). This design enables effective and visually imperceptible identity removal while preserving realism.

    \item Extensive experiments demonstrate that ID-Eraser achieves strong robustness and generalization across diverse face swapping models, cross-domain datasets, and even commercial black-box APIs. Our method provides reliable source-identity protection under real-world deployment 
    conditions.
\end{itemize}

\section{Related Work}
\label{sec:related-work}
\subsection{Deepfake Face Swapping}

Face swapping aims to replace the identity of a target face with that of a source while preserving the target’s pose, expression, and background. 
Early Deepfake systems adopted identity-specific autoencoders that trained a separate encoder–decoder pair for each identity and therefore suffered from limited scalability.
To enable generalizable identity transfer, recent methods extract a compact identity representation from the source face and inject it into the target image during synthesis. 
SimSwap~\cite{simswap} integrates identity embeddings through adaptive instance normalization, while InfoSwap~\cite{InfoSwap} enhances disentanglement via information bottlenecks. 
RAFSwap~\cite{RAFSwap} introduces region-aware identity tokens to refine local facial details, and BlendFace~\cite{blendface} redesigns the identity encoder for higher fidelity. 
E4S~\cite{E4S} adopts StyleGAN-based regional inversion for fine-grained identity manipulation, and FaceAdapter~\cite{Face-Adapter} leverages lightweight adapters to achieve diffusion-based identity transfer.  
More recently, CanonSwap~\cite{canonswap} performs identity modulation in a canonical space to improve consistency in video face swapping.

Although these models differ in architecture and implementation, most follow a similar high-level paradigm: 
identity features are first extracted from the source face and then injected into the target synthesis pathway. 
This pipeline has become the dominant design for modern face swapping systems, enabling stable identity manipulation across diverse input conditions.

%-------------------------------------------------------------------------

\begin{figure*}[h]
  \centering
  \includegraphics[width=\textwidth,trim=1 1 1 1,clip]{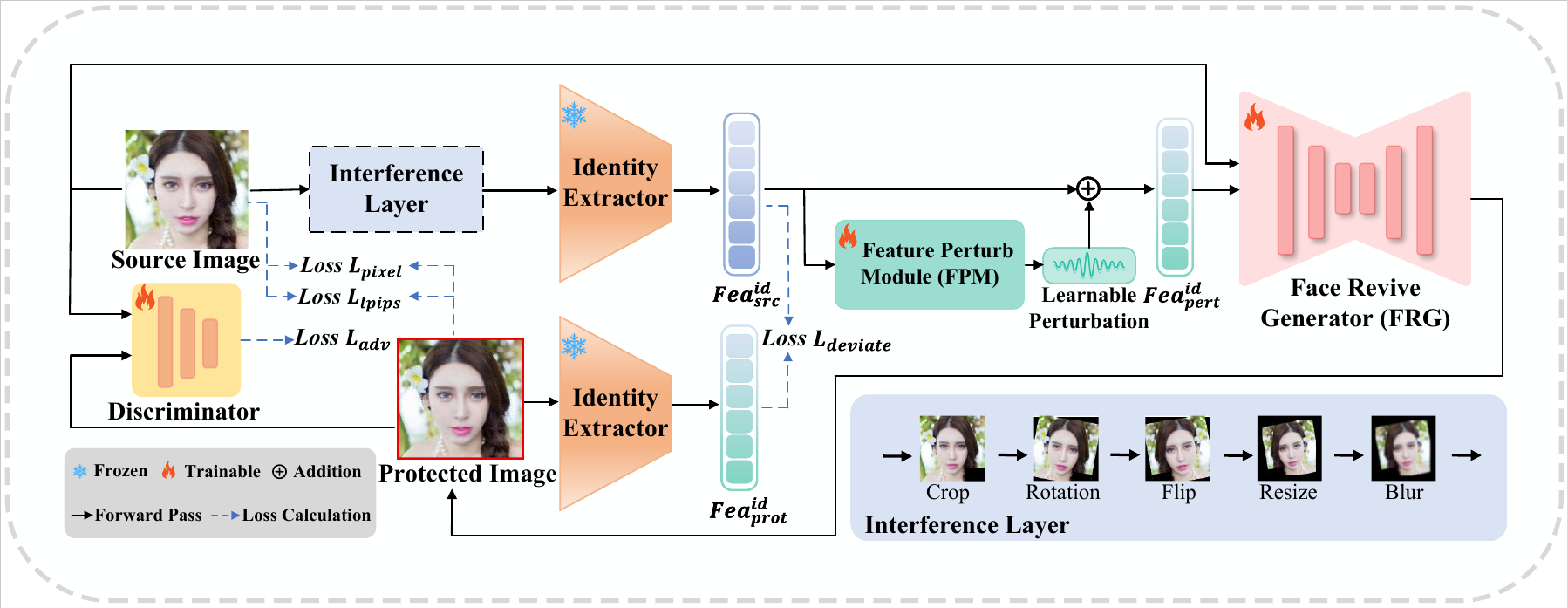} % 建议使用 \textwidth 使图片横跨双栏
  \caption{Overview of the proposed ID-Eraser framework. 
It comprises an Feature Perturbation Module (FPM) and a Face Revive Generator (FRG). 
FPM injects perturbations into the latent embedding extracted by a pre-trained recognizer, while FRG reconstructs a realistic protection image that preserves visual fidelity and removes identity cues.}
  \label{fig:pic2}
\end{figure*}

\subsection{Proactive Defense Against Deepfakes}

A growing line of research explores proactive defenses that modify the input image to prevent malicious manipulation. 
Most existing approaches operate in the pixel domain by injecting imperceptible perturbations to disrupt the processing pipeline of Deepfake models. 
However, the majority of these defenses are primarily designed for attribute manipulation or facial reenactment rather than for face swapping. 
Initiative Defense~\cite{initiative} introduces adversarial noise to mislead reenactment models, 
while Anti-Forgery~\cite{anti} embeds subtle perturbations to degrade the visual quality of manipulated attributes. 
CMUA-Watermark~\cite{cmua} improves cross-model transferability by jointly optimizing perturbations over multiple manipulation networks, 
and DF-RAP~\cite{DF-rap} enhances robustness against common image transformations such as compression and resizing.
Recent work also includes defenses specifically designed for face swapping.  
TCA-GAN~\cite{restricted} performs a restricted black-box attack by perturbing the target face image to degrade the swapped result.  
However, it only protects the particular target image being perturbed; once a different target is chosen, the source identity can still be extracted and misused. 
Thus, such methods provide target-side protection only and cannot prevent attackers from exploiting the source face to generate new identity transfers.

Despite their progress, pixel-domain defenses face a fundamental limitation: 
modern face swapping pipelines increasingly rely on robust identity encoders whose high-level feature representations are largely invariant to small pixel perturbations. 
As a result, pixel-level noise is often neutralized during feature extraction, making these approaches insufficient for blocking identity transfer in contemporary face swapping systems.

\section{Methodology}
\label{sec:methodology}

\subsection{Motivation}

Although modern face swapping systems differ in specific architectures, their core components exhibit a highly consistent design pattern. 
As shown in Table~\ref{tab:methods_ID_Module}, almost all recent face swapping methods include an independent identity extraction module, such as the identity injection module (IIM) in SimSwap, 
the region-aware identity tokenizer (RAT) in RAFSwap, or the identity encoder used in CanonSwap. 
These modules distill the source face into a compact and discriminative identity feature vector, which essentially determines whose identity will be transferred during the synthesis process. 
Therefore, the identity embedding space becomes the key channel through which identity information propagates in the entire face swapping pipeline.

This observation reveals the fundamental challenge faced by existing proactive defense approaches. 
Most current defenses introduce perturbations only in the pixel domain, yet the identity extractors used in modern face swapping models are highly robust to such minor pixel variations. 
They map faces onto a stable high-level semantic embedding manifold, causing pixel-level perturbations to be largely neutralized during identity feature extraction. 
As a result, the extracted identity vector remains almost identical to that of the original image. 
Such defenses may interfere with local attribute editing, but they are insufficient to disrupt the holistic identity representation required for face swapping, making it difficult to effectively block identity transfer.

To fundamentally prevent identity misuse, a defense must operate in the feature space rather than the image space. 
The core idea of our ID-Eraser is to directly manipulate the identity embedding in the latent space, breaking the transferability of identity features and preventing downstream generators from reconstructing a consistent identity.

%-----------------------------------------------------------
\subsection{Overall Framework}
ID-Eraser consists of two collaborative modules: (1) a Feature Perturbation Module (FPM), which injects subtle yet effective perturbations into the identity embeddings, and (2) a Face Revive Generator (FRG), which reconstructs visually natural images from the perturbed features. 

As illustrated in Figure~\ref{fig:pic2}, Given a source image $I_s$, we first employ a pretrained identity extractor ArcFace~\cite{arcface} to obtain its identity embedding $Fea^{id}_{src}$. The FPM then injects a learnable perturbation to generate the perturbed identity representation:

\begin{equation}
  Fea^{id}_{pert} = FPM(Fea^{id}_{src}).
  \label{eq:e1}
\end{equation}
Next, the FRG reconstructs a visually realistic yet identity-erased image $I_p$ based on the source image and perturbed identity:
\begin{equation}
  I_p = FRG(I_s, Fea^{id}_{pert}).
  \label{eq:e2}
\end{equation}
The protected image $I_p$ remains visually indistinguishable from $I_s$ but prevents face swapping models from correctly transferring the source identity. 

\begin{table}[h]
\caption{Identity extraction modules adopted in representative Deepfake face swapping models. 
The common presence of such modules highlights the feasibility of identity-level perturbation.}

\centering
\begin{tabular}{lcc}
\toprule
\textbf{Method} & \textbf{Publication} & \textbf{ID Extract Module}    \\
\midrule
FaceShifter~\cite{faceshifter} &-  & ArcFace \\
SimSwap~\cite{simswap}     &MM2020  & IIM \\
InfoSwap~\cite{InfoSwap}    &CVPR2021  & IIB \\
RAFSwap~\cite{RAFSwap}     &CVPR2022  & RAT  \\
BlendFace~\cite{blendface}   &ICCV2023  & BlendFace  \\
DiffSwap~\cite{diffswap}    &CVPR2023  & IE \\
E4S~\cite{E4S}         &CVPR2023  & RGI \\
%Face-Adapter~\cite{Face-Adapter}&ECCV2024  & IE \\
CanonSwap~\cite{canonswap}   &ICCV2025  & ID-Encoder \\
\bottomrule
\end{tabular}
\label{tab:methods_ID_Module}
\end{table}

To increase robustness and transferability under real-world manipulations and black-box swap models, we integrate an Interference Layer $T(\cdot)$ into the forward pass during training. The Interference Layer is implemented as a stochastic, differentiable augmentation pipeline composed of compositions of common image transforms:
\begin{equation}
  T(\cdot) \in \{\text{Crop}, \text{Rotation}, \text{Flip}, \text{Resize}, \text{Blur}\},
  \label{eq:e3}
\end{equation}
applied with randomized parameters. At each training iteration we sample a transformation $T$ from this distribution and apply it to inputs used for identity extraction and loss calculation.

%--------------------------------------------------------------
\subsection{Feature Perturbation Module (FPM)}
The FPM is designed to inject learnable perturbations into the latent identity space while keeping the perturbation imperceptible in the visual domain.  
Let the source identity embedding be denoted as $Fea^{id}_{src} \in \mathbb{R}^{d_f}$, where $d_f$ represents the dimensionality of the identity feature space.  
FPM employs a two-layer multilayer perceptron (MLP) to learn a bounded perturbation vector $\Delta f \in \mathbb{R}^{d_f}$.
Specifically, the perturbation is generated as:
\begin{equation}
  \Delta f = \tanh\big(w_2(\mathrm{ReLU}(w_1 Fea^{id}_{src} + b_1)) + b_2\big),
  \label{eq:e4}
\end{equation}
where $w_1 \in \mathbb{R}^{d_h \times d_f}$ and $w_2 \in \mathbb{R}^{d_f \times d_h}$ are learnable projection matrices, and $b_1, b_2$ are bias terms.  
The hidden dimension $d_h$ controls the transformation capacity of the perturbation network.  
The $\mathrm{ReLU}(\cdot)$ introduces nonlinearity, while the $\tanh(\cdot)$ constrains the perturbation range within $[-1, 1]$ to ensure stability and prevent over-distortion.

The final perturbed identity embedding is then defined as:
\begin{equation}
  Fea^{id}_{pert} = Fea^{id}_{src} + \alpha\Delta f,
  \label{eq:e5}
\end{equation}
where $\alpha$ is a scaling coefficient %(empirically set to $\alpha=0.05$) 
that controls the perturbation magnitude.  
Through joint optimization with the generator, FPM learns to maximize the semantic deviation between $Fea^{id}_{prot}$ and $Fea^{id}_{src}$ while maintaining the visual realism of the reconstructed image.

%------------------------------------------------------------
\subsection{Face Revive Generator (FRG)}

To reconstruct the protected image from the perturbed identity features, we employ a lightweight Swin-UNet~\cite{swinunet} architecture. This reduced version retains Swin-UNet’s hierarchical encoder–decoder design and skip connections, while simplifying the Transformer blocks for efficient image reconstruction in our setting.

The FRG consists of three main components:

\textbf{(1) Encoder.}
The encoder extracts hierarchical visual features through two stages of convolutional
downsampling, followed by a Swin Transformer block~\cite{swin-transformer} at each stage. This combination
captures both local textures and global contextual information while progressively 
reducing spatial resolution. Skip connections from each encoder stage are preserved for 
later fusion in the decoder.

\textbf{(2) Bottleneck.}
The bottleneck serves as the intermediate latent representation between the encoder and decoder. We denote this feature map as $b \in \mathbb{R}^{C_b \times H_b \times W_b}$, where $C_b$ is the channel dimension and $H_b, W_b$ are the spatial resolutions. The perturbed identity embedding $F^{id}_{pert}$ is first projected through two learnable linear layers to obtain $\tilde{f}p = w_4(\mathrm{ReLU}(w_3 F^{id}{pert}))$, where $w_3$ and $w_4$ are learnable projection matrices. The projected feature $\tilde{f}_p$ is then reshaped and broadcast across spatial dimensions before being additively fused into the bottleneck feature as $b = b + \tilde{f}_p$. This enables the perturbed identity information to modulate the global latent representation prior to decoding, ensuring that the reconstructed protection image removes identity cues while preserving visual realism.

\textbf{(3) Decoder.}
The decoder reconstructs the protected face by gradually upsampling the fused 
bottleneck feature using transposed convolutions, while integrating the skip-connected 
encoder features to restore high-frequency spatial details. A final Swin Transformer 
block further refines the decoded representation, and a $\tanh(\cdot)$ activation is 
applied at the output layer to produce the final protected image in normalized 
intensity range.

Through this design, FRG effectively integrates the perturbed identity signal into the latent feature space while maintaining natural image appearance, thus producing protection images that are visually indistinguishable from the source but semantically identity-erased.

%--------------------------------------------------
\subsection{Objective Functions}

\subsubsection{Identity Deviate Loss $L_{\text{deviate}}$}

The identity deviate loss measures the cosine similarity between the source identity feature $Fea^{id}_{src}$ and the protected identity feature $Fea^{id}_{prot}$. The goal is to minimize similarity, ensuring that the protected identity deviates 
significantly from the original source:
\begin{equation}
    % L_{\text{deviate}} = \frac{\mathbf{Fea^{id}_{src}} \cdot \mathbf{Fea^{id}_{prot}}}{\|\mathbf{Fea^{id}_{src}}\| \, \|\mathbf{Fea^{id}_{prot}}\|}.
    L_{\text{deviate}} = \frac{Fea^{id}_{src} \cdot Fea^{id}_{prot}}{\|Fea^{id}_{src}\| \, \|Fea^{id}_{prot}\|}.
\label{eq:e9}
\end{equation}

By enlarging this deviation, the model effectively erases identity information, preventing face swapping models from recognizing the true source identity.

\subsubsection{Pixel Loss $L_{\text{pixel}}$}

The pixel loss ensures that the generated protection image ${I}_p$ remains visually consistent with the source image $I_s$ at the pixel level.  
We employ the MSE to quantify the pixel-wise reconstruction difference:
\begin{equation}
   \mathcal{L}_{\text{pixel}} = \| I_s - I_p \|_2.
\label{eq:e10}
\end{equation}

This loss term helps maintain low-level fidelity and reduces visual artifacts in the generated image.

\subsubsection{Perceptual Loss $L_{\text{lpips}}$}

To preserve perceptual similarity in the high-level feature space, we adopt the \textit{Learned Perceptual Image Patch Similarity} (LPIPS) metric~\cite{LPIPS}. 
LPIPS measures the perceptual distance between the protected image and the source image based on deep features extracted from a pretrained VGG network:
\begin{equation}
L_{\text{lpips}} = \frac{1}{N} \sum_{i=1}^{N} \left\| \phi_i(I_s) - \phi_i(I_p) \right\|_2^2,
\label{eq:e11}
\end{equation}
where $\phi_i(\cdot)$ denotes the feature representation from the $i$-th layer of the pretrained VGG model. 
This loss captures perceptually meaningful differences and encourages the generated protection image to maintain visual realism while preserving high-level semantic consistency.

\subsubsection{Adversarial Loss $L_{\text{adv}}$}

To further improve the realism and visual quality of the protection image, 
we adopt the Least-Squares GAN (LSGAN)~\cite{LSGAN} objective, 
which stabilizes adversarial training and encourages realistic texture generation. 
The adversarial losses for the generator and discriminator are defined as:
\[
L_{adv} = 
\mathbb{E}_{\hat{I} \sim P_{\text{gen}}} \big[(D(I_p) - 1)^2 \big],
\]
\begin{equation}
L_{dis} = 
\mathbb{E}_{I \sim P_{\text{src}}} \big[(D(I_s) - 1)^2 \big] + 
\mathbb{E}_{\hat{I} \sim P_{\text{gen}}} \big[D({I_p})^2 \big].
\label{eq:e13}
\end{equation}
This objective provides smoother gradients than the standard cross-entropy GAN loss,
leading to more stable convergence in our generator training.

By combining these loss functions, ID-Eraser achieves a balanced optimization objective that ensures the generated protection images are both perceptually natural and resistant to Deepfake face swapping attacks.

The overall objective of ID-Eraser is formulated as a weighted combination of multiple loss terms:
\begin{equation}
    L_{\text{total}} = \lambda_{\text{a}} L_{\text{adv}} + 
    \lambda_{\text{p}} L_{\text{pixel}} + 
    \lambda_{\text{l}} L_{\text{lpips}} + 
    \lambda_{\text{d}} L_{\text{deviate}},
\label{eq:e13}
\end{equation}
where $\lambda_{\text{a}}$, $\lambda_{\text{p}}$, $\lambda_{\text{l}}$, and $\lambda_{\text{d}}$ are balancing coefficients that control the contribution of each loss term during training.

\begin{table*}[h]
\caption{Comparison of identity matching performance under various perturbation algorithms. Acc5 and Acc1 represent Top-5 and Top-1 accuracy, respectively. 
The downward arrows ($\downarrow$) indicate that lower values are better for attack scenarios. ``Clean" denotes images without any processing. Best performance marked in \textbf{bold}.}
%\small
\centering
\label{tab:identity erase}
\begin{tabular}{lcccccccccccc}
\toprule
%\multirow{3}{*}{Methods} & 
\multicolumn{1}{l}{Methods} &
\multicolumn{2}{c}{Clean} & 
\multicolumn{2}{c}{Initiative~\cite{initiative}} & 
\multicolumn{2}{c}{CMUA~\cite{cmua}} & 
\multicolumn{2}{c}{DF-RAP~\cite{DF-rap}} & 
\multicolumn{2}{c}{NullSwap~\cite{nullswap}} & 
\multicolumn{2}{c}{ID-Eraser(Ours)} \\
\midrule
Metrics& Acc5$\downarrow$ & Acc1$\downarrow$& Acc5$\downarrow$ & Acc1$\downarrow$  &Acc5$\downarrow$ & Acc1$\downarrow$ & Acc5$\downarrow$ & Acc1$\downarrow$ & Acc5$\downarrow$ & Acc1$\downarrow$ & Acc5$\downarrow$ & Acc1$\downarrow$ \\
\midrule
ArcFace~\cite{arcface}&0.952 &0.915 & 0.945 & 0.904 & 0.951 & 0.912 & 0.937 & 0.896 & 0.771 & 0.628 & \textbf{0.516} & \textbf{0.342} \\
FaceNet~\cite{facenet}&0.951 &0.917 & 0.935 & 0.884 & 0.950 & 0.914 & 0.945 & 0.901 & 0.740 & 0.590 & \textbf{0.359} & \textbf{0.192} \\
VGGFace~\cite{vggface}&0.931 &0.871 & 0.920 & 0.850 & 0.927 & 0.869 & 0.916 & 0.842 & 0.674 & 0.529 & \textbf{0.269} & \textbf{0.136} \\
SFace~\cite{sface}    &0.883 &0.818 & 0.874 & 0.815 & 0.876 & 0.809 & 0.869 & 0.797 & \textbf{0.658} & \textbf{0.513} & 0.703 & 0.532 \\
\midrule
Average         &0.929&0.880 & 0.919 & 0.863 & 0.926 & 0.876 & 0.917 & 0.859 & 0.711 & 0.565 & \textbf{0.461} & \textbf{0.301} \\
\bottomrule
\end{tabular}
\vspace{0.2cm}
\begin{minipage}{\linewidth}
%\small
\end{minipage}
\end{table*}
\section{Experiments}
\label{sec:experiments}

\subsection{Implementation Details}

\noindent \textbf{Dataset.} In our experiments, we adopt the widely used CelebA-HQ~\cite{celeba-HQ}, which contains 30,000 high-quality face images of 6,217 unique identities. Following the official split, we use it for both training and testing. For face swapping defense experiments, we randomly select target images from the FFHQ~\cite{FFHQ} to simulate a realistic restricted black-box scenario where the target faces are unknown. FFHQ contains 70,000 high-resolution face images with greater diversity in age, gender, skin tone, and background. 
Source images are drawn from the CelebA-HQ test set and are processed using the defense methods under evaluation. All images are preprocessed with face detection, alignment, and normalization, and resized to $256 \times 256$ resolution.

\noindent \textbf{Training Details.} During training, the FPM, FRG, and Discriminator are jointly optimized. The identity extractor uses a pre-trained ArcFace, whose parameters remain frozen during training. We employ the Adam~\cite{adam} optimizer with $\beta_1 = 0.5$ and $\beta_2 = 0.999$. The generator’s learning rate is set to $2\times10^{-4}$, while the perturbation module’s learning rate is halved. Loss weights are empirically set to $\lambda_a = 0.2$, $\lambda_p = 0.5$, $\lambda_l = 1.0$, and $\lambda_d = 0.15$. To ensure training stability, the discriminator is updated once every two iterations. The model is implemented in PyTorch and trained on a single NVIDIA A100 GPU for 100 epochs with a batch size of 32.

\begin{figure}[h]
  \centering
  \includegraphics[width=0.999999\linewidth]{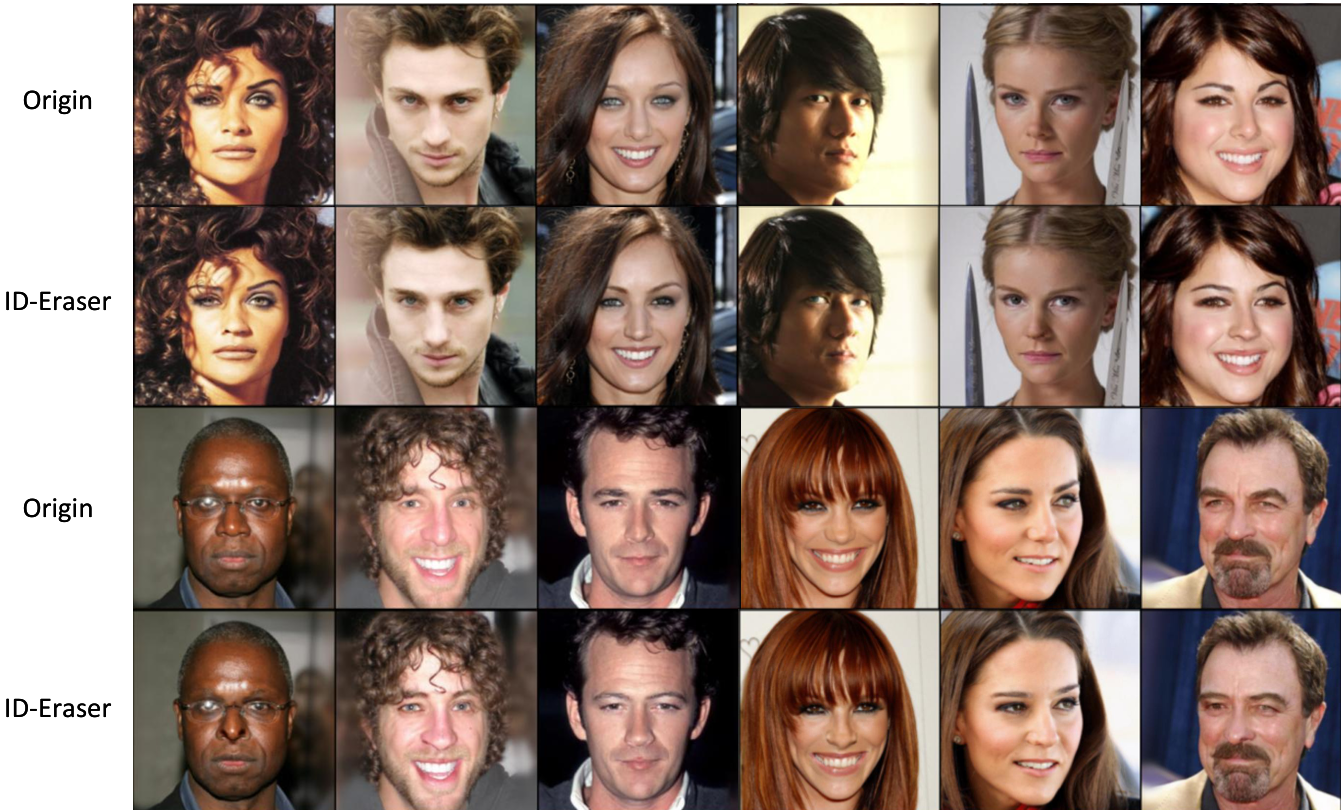} % 推荐使用\linewidth控制宽度
  \caption{Visualization of image visual quality, where ``Origin" denotes images without any processing.}
  \label{fig:pic3}
\end{figure}

\subsection{Visual Quality Evaluation}
\label{subsec:quantitative}
To further examine the perceptual realism of the protected images, we additionally provide qualitative visualizations in figure~\ref{fig:pic3}. As shown in the figure, the images generated by ID-Eraser remain highly consistent with the originals in terms of texture, color distribution, and global facial structure. No noticeable artifacts or distortions are introduced, and human observers can scarcely distinguish the protected images from their unprocessed counterparts.

Quantitative results in table~\ref{tab:visual quality} align with these visual observations. Although ID-Eraser reports a slightly lower PSNR (32.10) compared with several pixel-level perturbation methods, this reduction does not lead to perceptible degradation in visual quality. Pixel-oriented defenses often optimize for high PSNR by minimizing pixel-wise deviation, yet they tend to introduce high-frequency noise patterns that are visible upon close inspection. In contrast, ID-Eraser perturbs the identity representation in latent space and reconstructs the image through a generative model, yielding protection results that appear visually natural despite a lower pixel-wise similarity score.

This is further supported by the superior structural and perceptual metrics obtained by our method. ID-Eraser achieves the best SSIM (0.9565) and the lowest LPIPS (0.020), indicating that high-level perceptual similarity is well preserved. The lowest FID (1.64) also confirms that the generated images lie closest to the natural image distribution. Together with the qualitative results in figure~\ref{fig:pic3}, these observations demonstrate that ID-Eraser maintains excellent perceptual fidelity while embedding strong identity-erasing perturbations.

\begin{table}[h]
\caption{Visual quality evaluation of perturbed images. Best performance marked in \textbf{bold}.}
\centering
\begin{tabular}{lcccc}
\toprule
\textbf{Methods} & \textbf{FID$\downarrow$}& \textbf{PSNR$\uparrow$} & \textbf{SSIM$\uparrow$} & \textbf{LPIPS$\downarrow$} \\
\midrule
Initiative~\cite{initiative}   & 1.96 & \textbf{37.49} & 0.9526 & 0.0352 \\
Anti-Forgery~\cite{anti} & 3.44 & 36.85 & 0.9330 & 0.0396 \\
CMUA~\cite{cmua}         & 4.40 & 34.29 & 0.9152 & 0.0856 \\
DF-RAP~\cite{DF-rap}       & 8.34 & 36.72 & 0.9229 & 0.0623 \\
ID-Eraser(Ours) & \textbf{1.64} & 32.10 & \textbf{0.9565} & \textbf{0.0200} \\
\bottomrule
\end{tabular}
\label{tab:visual quality}
\end{table}

\subsection{Identity Erasure Performance}
\label{subsec:4.3}
To quantitatively evaluate the effectiveness of ID-Eraser in removing identity information, we conduct experiments on four mainstream face recognition models: ArcFace, FaceNet, VGGFace, and SFace. Here, the Top-1 and Top-5 identification accuracies represent the success rates at which the model matches the protected image to the most similar identity within the top 1 or top 5 candidates in the database, where lower values indicate stronger identity erasure. Notably, most pixel-level perturbation–based Deepfake defense methods show only marginal differences from the Clean baseline, indicating that they fail to substantially weaken identity features and therefore lack true identity-hiding capability. As shown in table~\ref{tab:identity erase}, ID-Eraser achieves the lowest identification accuracies across nearly all recognition models, demonstrating its strong capability in suppressing identifiable facial features.
\begin{table*}[h]
\centering
\caption{Identity similarity between swapped results generated from perturbed images and those generated from clean images across various face swapping models. Lower values indicate better defense performance. Best results are marked in \textbf{bold}.}
\label{tab:face_swapping_comparison}
\begin{tabular}{l*{10}{c}}
\toprule
\multicolumn{1}{l}{Methods}  & 
\multicolumn{2}{c}{Initiative~\cite{initiative}} & 
\multicolumn{2}{c}{Anti-Forgery~\cite{anti}} & 
\multicolumn{2}{c}{CMUA~\cite{cmua}} & 
\multicolumn{2}{c}{DF-RAP~\cite{DF-rap}} & 
\multicolumn{2}{c}{ID-Eraser (Ours)} \\
\midrule
Metrics  & ArcF.$\downarrow$ & VGGF.$\downarrow$ & ArcF.$\downarrow$ &    VGGF.$\downarrow$ & ArcF.$\downarrow$& VGGF.$\downarrow$& ArcF.$\downarrow$& VGGF.$\downarrow$& ArcF.$\downarrow$& VGGF.$\downarrow$ \\
\midrule
SimSwap~\cite{simswap} & 0.955 & 0.953 & 0.951 & 0.949 & 0.954 & 0.952 & 0.549 & 0.519 & \textbf{0.419} & \textbf{0.359} \\
InfoSwap~\cite{InfoSwap} & 0.953 & 0.952 & 0.953 & 0.950 & 0.956 & 0.955 & 0.910 & 0.902 & \textbf{0.540} & \textbf{0.453} \\
E4S~\cite{E4S} & 0.932 & 0.930 & 0.928 & 0.919 & 0.919 & 0.915 & 0.876 & 0.866 & \textbf{0.625} & \textbf{0.500} \\
BlendFace~\cite{blendface} & 0.960 & 0.959 & 0.963 & 0.951 & 0.962 & 0.960 & 0.910 & 0.903 & \textbf{0.602} & \textbf{0.541} \\
UniFace~\cite{uniface} & 0.979 & 0.980 & 0.978 & 0.978 & 0.979 & 0.980 & 0.932 & 0.925 & \textbf{0.547} & \textbf{0.451} \\
\midrule
Average & 0.956 & 0.955 & 0.955 & 0.949 & 0.954 & 0.952 & 0.835 & 0.823 & \textbf{0.547} & \textbf{0.461} \\
\bottomrule
\end{tabular}
\end{table*}

Among existing methods, NullSwap shows relatively strong suppression performance, yet it is still notably weaker than ID-Eraser. For instance, on ArcFace, NullSwap achieves a Top-1 accuracy of 0.628, whereas ID-Eraser further reduces it to 0.342. A similar pattern appears on FaceNet, where the Top-1 accuracy drops from 0.590 to 0.192, and on VGGFace, where it drops from 0.529 to 0.136, indicating that ID-Eraser can more effectively disrupt the discriminative identity representations extracted by different recognition models. On SFace, although NullSwap demonstrates greater robustness due to its architectural characteristics, ID-Eraser still maintains competitive suppression performance without significant degradation.

The averaged results across the four recognition models further validate this observation. ID-Eraser achieves an average Top-1 accuracy of 0.300 and an average Top-5 accuracy of 0.461, both of which are the lowest among all compared methods. This suggests that, compared with traditional pixel-level perturbation approaches, ID-Eraser fundamentally disrupts the identity embedding extraction process of modern recognition systems by directly perturbing identity representations in latent space.

\begin{figure}[h]
  \centering
  \includegraphics[width=\linewidth]{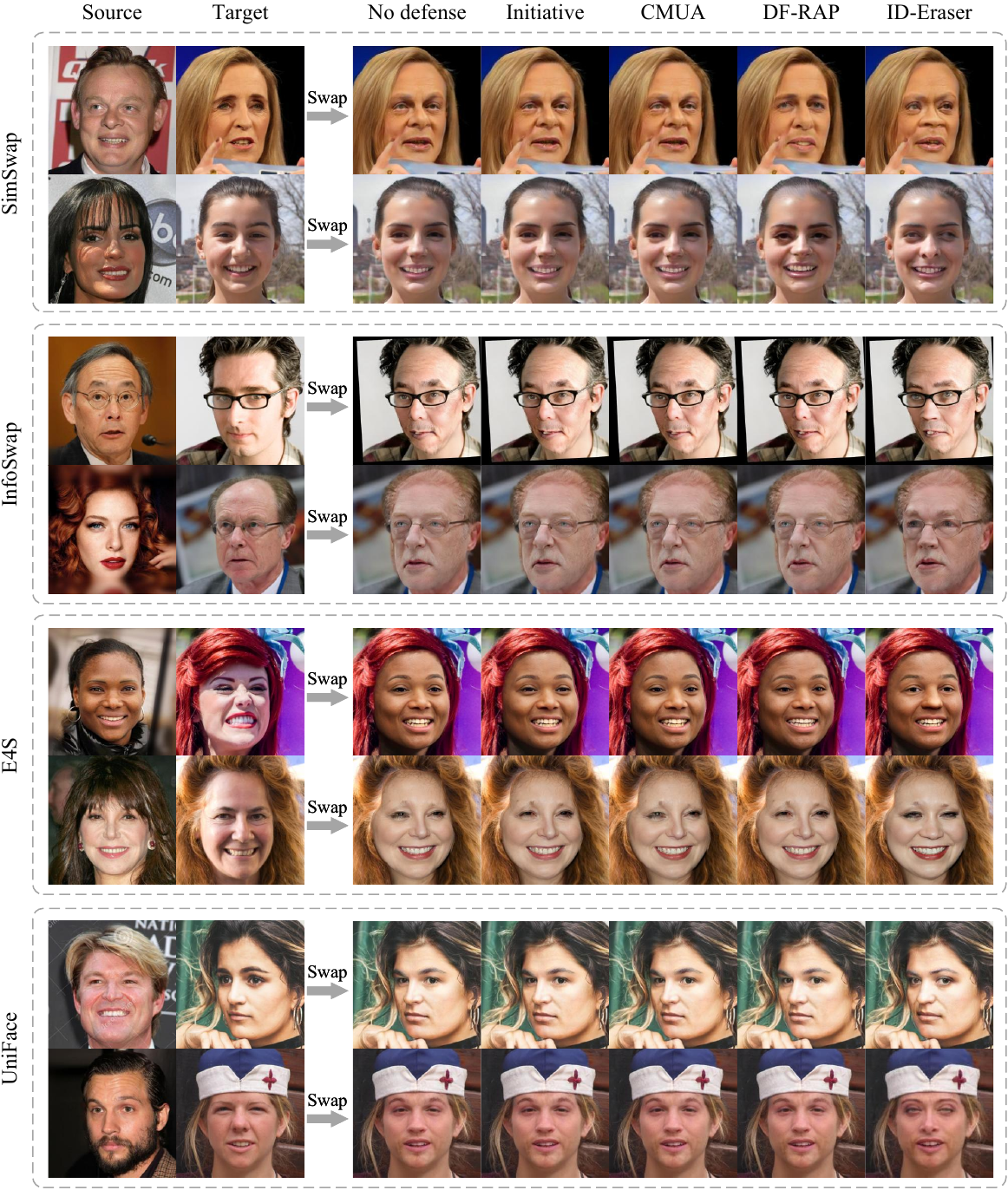} % 建议使用 \textwidth 使图片横跨双栏
  \caption{
Qualitative face swapping results under different defense methods. 
ID-Eraser effectively suppresses identity transfer across all face swapping models, producing faces that no longer retain the source identity.
}
  \label{fig:pic4}
\end{figure}

\subsection{Defense Against Face Swapping Models}
\label{subsec:face_swapping_defense}

To further evaluate the effectiveness of ID-Eraser against real-world face swapping, we conducted experiments on five representative models, including SimSwap~\cite{simswap}, InfoSwap~\cite{InfoSwap}, E4S~\cite{E4S}, BlendFace~\cite{blendface}, and UniFace~\cite{uniface}. Notably,  E4S, BlendFace, and UniFace use identity encoders that are completely different from ArcFace (or do not use ArcFace at all), which allows us to examine whether ID-Eraser’s effectiveness depends on a specific feature extractor. For each model, we compute the identity cosine similarity between swapped results ($I_{swap}^{o}$ and $I_{swap}^{p}$) generated from original inputs and protected inputs using ArcFace~\cite{arcface} and VGGFace~\cite{vggface}. Lower similarity indicates that the protected swapped result fails to preserve the source identity, reflecting stronger suppression of identity transfer.

As shown in table~\ref{tab:face_swapping_comparison}, for all swapping algorithms, the results produced by Initiative~\cite{initiative}, Anti-Forgery~\cite{anti}, and CMUA~\cite{cmua} remain highly similar to the corresponding clean-swapped outputs, with the ArcFace and VGGFace similarity remaining consistently around 0.95. This suggests that although these methods demonstrate certain robustness in simpler tasks such as attribute editing, the perturbations they introduce are insufficient to disrupt the identity propagation mechanism used by modern swapping models. In contrast, DF-RAP~\cite{DF-rap}, which was originally designed specifically for SimSwap, shows some capability in suppressing SimSwap’s identity transfer, reducing ArcFace and VGGFace similarity to 0.549 and 0.519, respectively. However, its perturbations lack cross-model generalization: on the remaining swapping models, the average similarity still remains as high as 0.900, indicating that DF-RAP is largely ineffective beyond its target architecture.

Our ID-Eraser achieves the lowest similarity scores across all swapping models, with average ArcFace and VGGFace similarities of 0.547 and 0.461, respectively, both lower than all baselines. These results demonstrate substantial and consistent suppression of identity transfer across diverse swapping frameworks.

The quantitative findings align well with the qualitative results shown in figure~\ref{fig:pic4}. The first two columns visualize the source and target images, and the ``No defense" column shows the swapped results from unprotected inputs, where all swapping models successfully transfer the source identity. The remaining pixel-level perturbation methods produce nearly identical visual results to the undefended case, further confirming that they fail to break the identity injection process. In contrast, ID-Eraser produces swapped faces that no longer resemble the source identity across all models, exhibiting clear identity deviation. This degradation in visual identity fidelity corresponds directly to the decreased cosine similarity measured by ArcFace and VGGFace, as reported in table~\ref{tab:face_swapping_comparison}, indicating that the identity information has been effectively removed.

Overall, both the quantitative and qualitative results show that ID-Eraser fundamentally disrupts the identity propagation mechanism on which face swapping models rely. By perturbing identity representations in latent space—rather than injecting pixel-level noise—ID-Eraser achieves strong robustness and cross-model generalization under black-box settings, effectively preventing the inappropriate transfer of source identities during face swapping.

\subsection{Effect of Perturbation Magnitude}
To provide a more transparent analysis of the trade-off between visual fidelity and identity confusion, we evaluate ID-Eraser under different perturbation magnitudes controlled by $\alpha$. As shown in table~\ref{tab:Perturb}, the identity-erasing performance remains relatively stable across a wide range of perturbation strengths, and larger perturbations do not lead to consistent performance gains on either face recognition or face-swapping evaluations. By contrast, visual quality degrades progressively as $\alpha$ increases, as evidenced by deteriorated FID and PSNR values and a pronounced increase in LPIPS when $\alpha=0.5$. This observation indicates that stronger perturbations do not necessarily produce better protection, but instead introduce a less favorable fidelity--protection trade-off. Based on this analysis, we set $\alpha = 0.05$ as the default choice in all remaining experiments.
\begin{table}[h]
\caption{Trade-off between visual fidelity and identity confusion under different perturbation magnitudes.}
% \tiny
\centering
\label{tab:Perturb}
% \resizebox{\hsize}{!}{
% \setlength{\tabcolsep}{12pt}{
\begin{tabular}{lcccc}
\toprule

% 第三行标题：指标行
Metrics& $\alpha=0.05$ & $\alpha=0.1$  & $\alpha=0.25$ & $\alpha=0.5$  \\
\midrule
FID↓             & \textbf{1.64} & 3.27 & 4.17 & 9.98 \\
PSNR↑            & \textbf{32.10} & 31.32 & 30.47 & 27.10 \\
LPIPS↓           & 0.020 & \textbf{0.019} & 0.022 & 0.063  \\
\midrule
Acc1-ArcFace~\cite{arcface}↓ & 0.342 & 0.351 & 0.346 & \textbf{0.270}  \\
Acc1-FaceNet~\cite{facenet}↓   & \textbf{0.192} & 0.226 & 0.246 & 0.225  \\
Acc1-VGGFace~\cite{vggface}↓   & \textbf{0.136} & 0.143 & 0.157 & 0.146  \\
\midrule
IDSim-SimSwap~\cite{simswap}↓  & \textbf{0.419} & 0.563 & 0.559 & 0.588 \\
IDSim-InfoSwap~\cite{InfoSwap}↓  & \textbf{0.540} & 0.623 & 0.609 & 0.646 \\
IDSim-E4S~\cite{E4S}↓  & \textbf{0.625} & 0.631 & 0.633 & 0.651 \\
IDSim-UniFace~\cite{uniface}↓  & \textbf{0.547} & 0.582 & 0.577 & 0.603 \\
\bottomrule
\end{tabular}

\end{table}

\begin{table*}[h]
\caption{Robustness evaluation under various image degradations. 
``None" indicates protected images without any additional degradation. Acc5 and Acc1 represent Top-5 and Top-1 accuracy, respectively. 
}

\centering
\label{tab:robustness}
\begin{tabular}{lcccccccccccc}
\toprule
%\multirow{3}{*}{Methods} & 
\multicolumn{1}{l}{Methods} &
\multicolumn{2}{c}{None} & 
\multicolumn{2}{c}{JPEG} & 
\multicolumn{2}{c}{Noise} & 
\multicolumn{2}{c}{Blur} & 
\multicolumn{2}{c}{Brightness} & 
\multicolumn{2}{c}{Contrast}  \\
\midrule
Metrics& Acc5$\downarrow$ & Acc1$\downarrow$& Acc5$\downarrow$ & Acc1$\downarrow$  &Acc5$\downarrow$ & Acc1$\downarrow$ & Acc5$\downarrow$ & Acc1$\downarrow$ & Acc5$\downarrow$ & Acc1$\downarrow$& Acc5$\downarrow$ & Acc1$\downarrow$  \\
\midrule
ArcFace~\cite{arcface}&0.516 &0.342 &0.514 &0.337 & 0.469 & 0.301 & 0.458 & 0.278 & 0.512 & 0.331 & 0.520 & 0.342 \\
FaceNet~\cite{facenet}&0.359 &0.192 &0.357 &0.189 & 0.347 & 0.183 & 0.288 & 0.141 & 0.360 & 0.192 & 0.367 & 0.200  \\
VGGFace~\cite{vggface}&0.269 &0.136 &0.265 &0.136 & 0.253 & 0.131 & 0.243 & 0.119 & 0.260 & 0.131 & 0.262 & 0.137  \\
SFace~\cite{sface}    &0.703 &0.532 &0.690 &0.515 & 0.671 & 0.499 & 0.669 & 0.486 & 0.695 & 0.519 & 0.696 & 0.519  \\
\midrule
Average         &0.461 &0.301 &0.457&0.294 & 0.435 & 0.279 & 0.415 & 0.256  & 0.457 & 0.293 & 0.461 & 0.300  \\
\bottomrule
\end{tabular}
\vspace{0.2cm}
\begin{minipage}{\linewidth}

\end{minipage}
\end{table*}

\subsection{Robustness Evaluation}

To further evaluate the robustness of the proposed ID-Eraser under real-world degradations, we conduct a series of stress tests on the protected images from the CelebA-HQ test set. Five common types of image degradations are considered, covering compression artifacts, noise corruption, resolution degradation, and photometric variations. To ensure a fair evaluation, the protected images are evenly divided across all parameter settings within each degradation type, such that each processed subset contains the same number of images.

\textbf{(1) JPEG Compression.}
We apply lossy JPEG encoding with quality factors \{95, 75, 50, 35, 20\} to simulate common resaving and media-upload pipelines.

\textbf{(2) Gaussian Noise.}
Additive Gaussian noise with standard deviations $\sigma \in \{2, 5, 10, 15, 25\}$ is introduced to mimic sensor noise and low-light capture conditions.

\textbf{(3) Gaussian Blur.}
Gaussian blur with kernel sizes $k \in \{3, 5, 7, 9, 11\}$ is used to emulate motion blur and resolution degradation.

\textbf{(4) Brightness Adjustment.}
Image brightness is scaled by factors \{1.05, 1.10, 1.15, 1.20, 1.30\} to reflect illumination and exposure variations.

\textbf{(5) Contrast Adjustment.}
Image contrast is adjusted using factors \{1.05, 1.10, 1.15, 1.20, 1.30\} to simulate dynamic-range changes and common post-processing effects.

\vspace{4pt}
\noindent
\textbf{Quantitative Results.}
Table~\ref{tab:robustness} summarizes the identity-matching performance under all degradation settings. Without degradation, the protected images achieve average accuracies of $\text{Acc5}=0.461$ and $\text{Acc1}=0.301$ across the four recognition models. Across all corruption types, the performance variation remains limited: $\text{Acc5}$ changes only within the range of $[0.415, 0.461]$, while $\text{Acc1}$ remains within $[0.256, 0.300]$. Notably, even under the most destructive distortions---Gaussian noise and Gaussian blur---$\text{Acc1}$ stays between $0.256$ and $0.279$, which is substantially lower than the accuracies of unprotected clean images (typically above 0.87, as reported in table~\ref{subsec:4.3}). These results demonstrate that the injected latent perturbation preserves its identity-erasing effect even when the image undergoes significant photometric or structural degradation.

Overall, the robustness evaluation shows that ID-Eraser maintains stable identity removal under compression, noise, blur, and illumination shifts, highlighting its strong reliability in practical deployment scenarios.

\subsection{Practical Effectiveness Evaluation}

To evaluate the practical effectiveness of ID-Eraser in real-world scenarios, we further tested it on two mainstream commercial face-verification APIs, Baidu and Tencent. Each API outputs a similarity score between two faces. Specifically, C-Score measures the similarity between the original source image $I_{s}$ and its corresponding swapped result $I_{swap}^{o}$, while P-Score measures the similarity between $I_{s}$ and $I_{swap}^{p}$ generated from ID-Eraser-protected sources.

As shown in table~\ref{tab:API}, ID-Eraser significantly reduces identity similarity in both commercial systems. 
The average similarity measured by the Baidu API decreases from 84.03 to 53.87, and that measured by the Tencent API drops from 76.48 to 35.61. 
These results indicate that commercial recognition systems can no longer confidently classify the protected swapped faces as belonging to the same identity, confirming that ID-Eraser effectively disrupts the identity propagation process exploited by face swapping models.

This result further highlights the practical security significance of our approach.
By lowering the matching confidence of commercial identity-verification services, ID-Eraser helps prevent Deepfake face swapping attacks from bypassing liveness detection and authentication mechanisms, while demonstrating outstanding robustness and defensive potential beyond standard academic benchmarks.
\begin{table}[h]
\caption{Evaluation on commercial face-verification APIs. By greatly reducing similarity, ID-Eraser enhances the reliability of real-world identity verification systems.}

\centering
\label{tab:API}
\begin{tabular}{lcccc}
\toprule
%\multirow{3}{*}{Methods} & 
\multicolumn{1}{l}{API} &
\multicolumn{2}{c}{Baidu} & 
\multicolumn{2}{c}{Tencent} \\
\midrule
% 第三行标题：指标行
Metrics& C-Score & P-Score  & C-Score & P-Score  \\
\midrule
SimSwap~\cite{simswap}   & 83.11 & \textbf{46.82} & 79.87 & \textbf{30.65} \\
InfoSwap~\cite{InfoSwap} & 87.37 & \textbf{60.70} & 83.11 & \textbf{43.16} \\
E4S~\cite{E4S}           & 78.44 & \textbf{51.59} & 65.87 & \textbf{33.20}  \\
BlendFace~\cite{blendface} & 82.53 & \textbf{52.21} & 71.61 & \textbf{31.94}  \\
UniFace~\cite{uniface}   & 88.73 & \textbf{58.02} & 81.94 & \textbf{39.09}  \\
\midrule
Average                  & 84.03 & \textbf{53.87} & 76.48 & \textbf{35.61} \\
\bottomrule
\end{tabular}
\vspace{0.2cm}
\begin{minipage}{\linewidth}
\small
\end{minipage}
\end{table}

\subsection{Ablation Study}

Although the overall structure of ID-Eraser cannot be decomposed into independent submodules, we further examine whether its performance depends on the identity feature extractor. Since the extractor defines the embedding space in which perturbations are applied, we replace the original ArcFace-based extractor with FaceNet and compare the results using two cosine-similarity metrics: Sim-ID (intra-identity similarity) and Sim-Pair (similarity between the original and protected images).
\begin{table}[h]
\caption{Ablation study on identity feature extractors. 
SimID and SimPair denote intra-identity and pairwise similarity, respectively.}
\centering
\label{tab:ablation}
\begin{tabular}{lcccc}
\toprule
%\multirow{3}{*}{Methods} & 
\multicolumn{1}{l}{ID Extractor} &
\multicolumn{2}{c}{FaceNet} & 
\multicolumn{2}{c}{ArcFace} \\
\midrule
% 第三行标题：指标行
Metrics& Sim-ID$\downarrow$ & Sim-Pair$\downarrow$  & Sim-ID$\downarrow$ & Sim-Pair$\downarrow$  \\
\midrule
ArcFace~\cite{arcface} & 0.427 & 0.539 & 0.380 & 0.491 \\
FaceNet~\cite{facenet} & 0.475 & 0.560 & 0.428 & 0.501  \\
VGGFace~\cite{vggface} & 0.373 & 0.427 & 0.321 & 0.362 \\
SFace~\cite{sface}     & 0.514 & 0.680 & 0.496 & 0.667 \\
\midrule
Average          & 0.447 & 0.551 & 0.406 & 0.505 \\
\bottomrule
\end{tabular}
\vspace{0.2cm}
\begin{minipage}{\linewidth}
\end{minipage}
\end{table}

As shown in table~\ref{tab:ablation}, the ArcFace configuration achieves lower SimID and SimPair across all recognition models, with averages of 0.406 and 0.505, respectively. This indicates that a more discriminative embedding space enables perturbations to more effectively disrupt identity consistency. In contrast, the weaker separability of FaceNet leads to higher residual similarity and thus limits the effectiveness of identity erasure.

These findings demonstrate that the performance of ID-Eraser is influenced not only by the perturbation mechanism itself but also by the discriminative power of the underlying identity extractor. A more discriminative feature space provides clearer semantic structure, enabling more targeted and robust identity removal.

\subsection{Cross-Dataset and Video-Scenario Evaluation}
\label{sec:rationale}

\subsubsection{Identity Erasure Evaluation}

To further evaluate the generalization of ID-Eraser beyond the training distribution, we extend our experiments from CelebA-HQ~\cite{celeba-HQ} to additional benchmarks covering both static image datasets and video-based scenarios. Specifically, we incorporate LFW~\cite{LFW} and VGGFace2~\cite{vggface2} as two widely used face image datasets, and FaceForensics++~\cite{FF++} as a representative benchmark for video-based face manipulation. LFW contains unconstrained real-world face images with substantial variations in illumination, pose, occlusion, and expression. VGGFace2 includes a much larger number of identities and exhibits stronger intra-class diversity, making it a more challenging benchmark for identity-related evaluation. In addition, FaceForensics++ provides a more realistic video-based scenario with richer variations in pose, expression, compression, and frame-level dynamics.

In this evaluation, we quantify identity removal using Sim-ID (intra-identity similarity).
The metric is computed as follows: for each protected image $I_{p}$ generated by ID-Eraser, we collect all other clean images belonging to the same identity, denoted as $\{I_{\text{clean}}^{(i)}\}$. We then compute the cosine similarity between their identity embeddings. SimID is defined as the average similarity within each identity:

\begin{equation}
\text{Sim-ID}(I_{p}) 
= \frac{1}{N} \sum_{i=1}^{N} 
\cos\!\left(
    f(I_{p}),\,
    f\big(I_{\text{clean}}^{(i)}\big)
\right),
\end{equation}
where $f(\cdot)$ extracts identity embeddings using ArcFace, FaceNet, VGGFace, or SFace.
A lower SimID indicates that samples of the same identity no longer cluster together after protection, demonstrating successful disruption of identity consistency.
\begin{table}[h]
\caption{Cross-dataset and cross-scenario identity erasure evaluation. Sim-ID quantifies intra-identity similarity, where lower is better. Results show that ID-Eraser generalizes effectively across static image datasets and the video-based FaceForensics++ benchmark.}
\centering
\label{tab:cross_erase}
\small
\begin{tabular}{lcccc}
\toprule
%\multirow{3}{*}{Methods} & 
\multicolumn{1}{l}{Dataset} &
\multicolumn{1}{c}{CelebA~\cite{celeba-HQ}} & 
\multicolumn{1}{c}{LFW~\cite{LFW}} & 
\multicolumn{1}{c}{VGGFace2~\cite{vggface2}}&
\multicolumn{1}{c}{FF++~\cite{FF++}}\\
\midrule
% 第三行标题：指标行
Metrics& Sim-ID$\downarrow$ & Sim-ID$\downarrow$  & Sim-ID$\downarrow$ & Sim-ID$\downarrow$\\
\midrule
ArcFace~\cite{arcface} & 0.381 & 0.491 & 0.565 & 0.444\\
FaceNet~\cite{facenet} & 0.429 & 0.571 & 0.652 & 0.520 \\
VGGFace~\cite{vggface} & 0.321 & 0.483 & 0.541 & 0.396 \\
SFace~\cite{sface}     & 0.497 & 0.587 & 0.588 & 0.509\\
\midrule
Average                & 0.407 & 0.533 & 0.587 & 0.467\\
\bottomrule
\end{tabular}
\vspace{0.2cm}
\begin{minipage}{\linewidth}
\end{minipage}
\end{table}

As shown in Table~\ref{tab:cross_erase}, ID-Eraser achieves an average Sim-ID of 0.407 on CelebA-HQ. On LFW, which involves more unconstrained acquisition conditions, the average Sim-ID increases moderately to 0.533, while on the larger and more diverse VGGFace2 dataset it reaches 0.587. Notably, on the video-based FaceForensics++ benchmark, ID-Eraser still maintains a relatively low average Sim-ID of 0.467. Despite substantial differences in pose, illumination, domain distribution, and temporal scenario between these benchmarks, the Sim-ID values remain consistently low relative to unprotected images. These results indicate that ID-Eraser effectively suppresses identity information not only across static image datasets, but also under video-based conditions.

Overall, the results demonstrate that ID-Eraser exhibits strong generalization across both datasets and scenarios. The latent-space perturbation does not rely on dataset-specific statistics and remains effective under both static and dynamic conditions, which is important for real-world deployment where protected content may originate from highly diverse sources.

\begin{table}[h]
\caption{Cross-dataset and cross-scenario image quality evaluation of protected images. The results show that ID-Eraser maintains high visual fidelity across static image datasets and the video-based FaceForensics++ benchmark.}
\centering
\label{tab:cross_quality}
\begin{tabular}{lcccc}
\toprule
%\multirow{3}{*}{Methods} & 
\multicolumn{1}{l}{Dataset} &
\multicolumn{1}{c}{CelebA~\cite{celeba-HQ}} & 
\multicolumn{1}{c}{LFW~\cite{LFW}} & 
\multicolumn{1}{c}{VGGFace2~\cite{vggface2}}&
\multicolumn{1}{c}{FF++~\cite{FF++}}\\

\midrule
FID$\downarrow$       & 1.642 & 1.506 & 2.317  &8.059\\
PSNR$\uparrow$      & 32.10 & 33.22 & 31.87  &32.41 \\
SSIM$\uparrow$      & 0.9565 & 0.9664 & 0.9573 &0.9676\\
LPIPS$\downarrow$     & 0.0200 & 0.0165 & 0.0276 &0.0168\\

\bottomrule
\end{tabular}
\vspace{0.2cm}
\begin{minipage}{\linewidth}
\end{minipage}
\end{table}

\subsubsection{Cross-Dataset Image Quality Evaluation}
To complement the identity erasure analysis, we further examine whether ID-Eraser maintains high perceptual quality across previously unseen datasets and video-based scenarios. Specifically, we evaluate protected images on LFW, VGGFace2, and FaceForensics++, all of which differ substantially from CelebA-HQ in acquisition conditions, pose variations, and data distributions. Maintaining high perceptual stability across such benchmarks is important, as it shows that identity removal is not achieved at the expense of noticeable visual degradation.

Table~\ref{tab:cross_quality} reports FID, PSNR, SSIM, and LPIPS across CelebA-HQ, LFW, VGGFace2, and FaceForensics++. Despite noticeable differences in imaging conditions and data distributions, ID-Eraser consistently produces visually natural outputs. On CelebA-HQ, the protected images achieve an FID of 1.642, a PSNR of 32.10, an SSIM of 0.9565, and an LPIPS of 0.0200. When applied to LFW and VGGFace2, which contain substantially more uncontrolled variations, the reconstruction quality remains comparably high. Notably, on FaceForensics++, ID-Eraser achieves a PSNR of 32.41, an SSIM of 0.9676, and an LPIPS of 0.0168, indicating that high perceptual similarity is preserved even under video-based conditions. Although the FID on FaceForensics++ increases to 8.059, the overall perceptual quality remains strong according to the full set of metrics.

These results show that ID-Eraser maintains stable visual fidelity not only under static domain shifts, but also in video-based scenarios. Combined with the identity erasure results in Table~\ref{tab:cross_erase}, this demonstrates that the method suppresses identity features effectively while preserving high-quality appearance across both static and dynamic settings.

\begin{table}[t]
\caption{Face swapping defense evaluation on the video-based FaceForensics++ benchmark. Identity similarity is measured between clean swapped results and protected swapped results. Lower values indicate stronger suppression of identity transfer.}
\centering
\label{tab:table3}
\begin{tabular}{lccc}
\toprule
\textbf{FS-Model} & \textbf{ArcFace-ID$\downarrow$} & \textbf{VGGFace-ID$\downarrow$} & \textbf{FaceNet-ID$\downarrow$} \\
\midrule
SimSwap~\cite{simswap} & 0.370 & 0.302 & 0.406 \\
UniFace~\cite{uniface} & 0.493 & 0.373 & 0.513 \\
InfoSwap~\cite{InfoSwap} & 0.485 & 0.397 & 0.499 \\
E4S~\cite{E4S} & 0.572 & 0.487 & 0.563 \\
\bottomrule
\end{tabular}
\end{table}

\subsubsection{Defense Against Face Swapping Models on Video Scenario}

To further evaluate the effectiveness of ID-Eraser in video-based face swapping scenarios, we conduct additional experiments on FaceForensics++, a widely used benchmark for video-based face manipulation. Compared with static image datasets, FaceForensics++ provides a more realistic setting with richer variations in pose, expression, compression, and frame-level dynamics. This makes it suitable for assessing whether ID-Eraser remains effective under dynamic conditions that are common in real-world Deepfake applications.

We evaluate ID-Eraser on FaceForensics++ using four representative face swapping models, namely SimSwap, UniFace, InfoSwap, and E4S. Following the same evaluation protocol as in the main paper, we compare the identity similarity between clean swapped results and protected swapped results using multiple face recognition models, including ArcFace, VGGFace, and FaceNet. Lower similarity indicates that the swapped results generated from protected inputs are less able to preserve the source identity, reflecting stronger suppression of identity transfer.

As shown in Table~\ref{tab:table3}, ID-Eraser consistently reduces identity similarity across all four face swapping models on FaceForensics++. For SimSwap, the identity similarity drops to 0.370, 0.302, and 0.406 under ArcFace, VGGFace, and FaceNet, respectively. For UniFace, the corresponding values are 0.493, 0.373, and 0.513. For InfoSwap, the similarity remains relatively low at 0.485, 0.397, and 0.499, while for E4S, the corresponding values are 0.572, 0.487, and 0.563. These results show that ID-Eraser remains effective in suppressing identity transfer even in video-based scenarios, where face swapping typically involves more complex variations and dynamic conditions than static image settings.

Overall, the results on FaceForensics++ demonstrate that ID-Eraser remains effective against diverse video-based face swapping models. This further confirms that the proposed latent-space identity perturbation generalizes beyond static image benchmarks and provides robust source-identity protection under more realistic dynamic conditions.

\section{Conclusion}
\label{sec:conclusion}

In this paper, we present ID-Eraser, a proactive defense that suppresses identity leakage in face swapping by perturbing identity embeddings in latent space. Unlike pixel-level perturbation approaches, ID-Eraser operates directly on semantic identity representations, achieving a balance between visual realism and protection effectiveness. Experiments show that it effectively erases identity features across diverse recognition and face swapping models. Future work will explore adaptive perturbation strategies and privacy-preserving extensions for broader real-world deployment.

\bibliographystyle{unsrt}
\bibliography{ref}
\end{document}